%% file: arxiv_main.tex
\documentclass[letterpaper, 10 pt, conference]{ieeeconf}  %

\usepackage{multirow}
\usepackage{tikz}
\input{header.tex}

\newcommand\copyrighttext{%
  \footnotesize \textcopyright 2025 IEEE. Personal use of this material is permitted. Permission from IEEE must be obtained for all other uses, in any current or future media, including reprinting/republishing this material for advertising or promotional purposes, creating new collective works, for resale or redistribution to servers or lists, or reuse of any copyrighted component of this work in other works. 
  DOI: \href{https://doi.org/10.1109/Humanoids65713.2025.11203050}{Humanoids65713.2025.11203050}}
\newcommand\copyrightnotice{%
\begin{tikzpicture}[remember picture,overlay]
\node[anchor=south,yshift=10pt] at (current page.south) {\fbox{\parbox{\dimexpr\textwidth-\fboxsep-\fboxrule\relax}{\copyrighttext}}};
\end{tikzpicture}%
}

\graphicspath{{media/}}

\IEEEoverridecommandlockouts                              %

\title{\LARGE \bf Adding internal audio sensing to internal vision enables\\ human-like in-hand fabric recognition with soft robotic fingertips}

\author{
Iris Andrussow$^{1,2}$\thanks{$^{1}$Haptic Intelligence Department, Max Planck Institute for Intelligent Systems, Stuttgart, Germany. \texttt{andrussow@is.mpg.de}, \texttt{kjk@is.mpg.de}}%
\thanks{$^{2}$Department for Distributed Intelligence / Autonomous Learning, University of Tübingen, Germany. \texttt{georg.martius@uni-tuebingen.de}},
Jans Solano$^{1}$,
Benjamin A. Richardson$^{1}$,
Georg Martius$^{2}$,
Katherine J. Kuchenbecker$^{1}$
}

\begin{document}

\maketitle

\copyrightnotice

\begin{abstract}
Distinguishing the feel of smooth silk from coarse cotton is a trivial everyday task for humans. When exploring such fabrics, fingertip skin senses both spatio-temporal force patterns and texture-induced vibrations that are integrated to form a haptic representation of the explored material.
It is challenging to reproduce this rich, dynamic perceptual capability in robots because tactile sensors typically cannot achieve both high spatial resolution and high temporal sampling rate.
In this work, we present a system that can sense both types of haptic information, and we investigate how each type influences robotic tactile perception of fabrics.
Our robotic hand's middle finger and thumb each feature a soft tactile sensor: one is the open-source Minsight sensor that uses an internal camera to measure fingertip deformation and force at 50 Hz, and the other is our new sensor Minsound that captures vibrations through an internal MEMS microphone with a bandwidth from 50\,Hz to 15\,kHz.
Inspired by the movements humans make to evaluate fabrics, our robot actively encloses and rubs folded fabric samples between its two sensitive fingers.
Our results test the influence of each sensing modality on overall classification performance, showing high utility for the audio-based sensor. Our transformer-based method achieves a maximum fabric classification accuracy of 97\% on a dataset of 20 common fabrics. Incorporating an external microphone away from Minsound increases our method's robustness in loud ambient noise conditions. To show that this audio-visual tactile sensing approach generalizes beyond the training data, we learn general representations of fabric stretchiness, thickness, and roughness. 
\end{abstract}

\section{Introduction}

Texture is a key object property that can be perceived by touch. It relates to both the material of the object and its surface structure, which are independent of the object's overall shape.
Human texture perception depends on two distinct modes that operate in largely different temporal frequency ranges \cite{weberSpatialTemporalCodes2013}. The low-frequency mode, sensitive up to approximately 50~Hz \cite{johansson2009coding}, captures skin deformation as local spatial patterns. In contrast, the high-frequency mode detects transient skin vibrations up to 1\,kHz \cite{shao2016wholehand}, such as those caused by slip or surface scanning, as temporal patterns. Together, these sensing modes create the perception of texture for humans \cite{LIEBER2022102621}; consequently, we believe both modes are essential to achieving human-like touch abilities in robotics.

\begin{figure}[t]
\begin{center}
   \includegraphics[width=\linewidth]{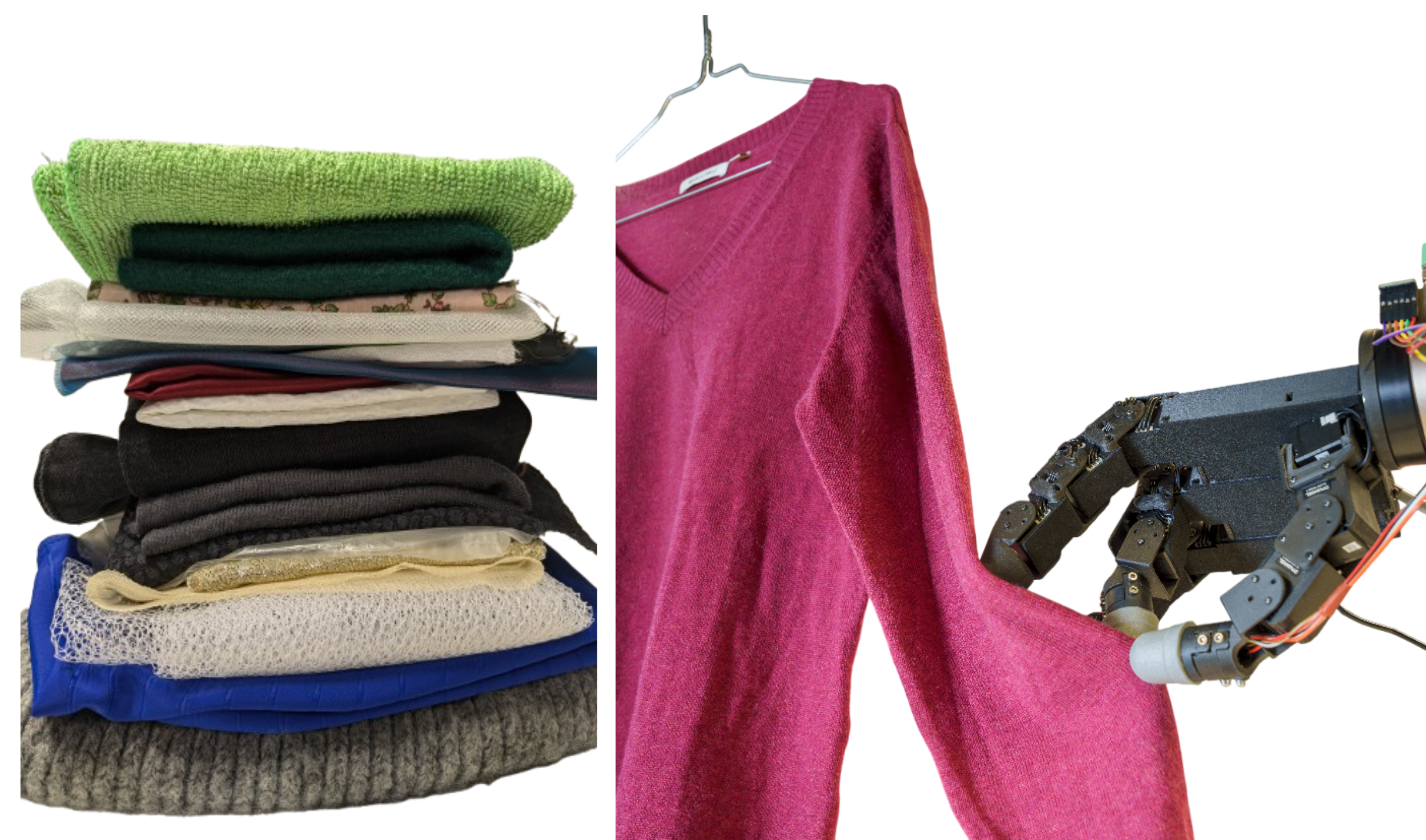}
    \caption{\textbf{Setup:} Recognizing common fabrics using multimodal input from vision- and audio-based fingertip sensors on a robot hand.
    }
    \label{fig:overview}
\end{center}
\end{figure}

The recent trend of using internal fingertip cameras for high-spatial-resolution tactile sensing has enabled robot touch to surpass the mode of human perception that has low temporal frequency and high spatial resolution. These sensors often provide exceptional spatial resolution but are limited to camera frame rates, which range from 30 up to about 300\,Hz. Yuan et al.~\cite{yuanActiveClothingMaterial2018b} have shown that in-hand fabric recognition is possible with high-resolution camera-based tactile sensors, but their approach depends on a thin and highly deformable outer layer of the sensor as well as a stationary pressing fingertip to capture the deformation patterns that show the fine details of the texture being touched. These delicate sensor layers are known to be susceptible to wear and tear during repeated contact interactions \cite{yuanActiveClothingMaterial2018b}.
As shown by Insight, fabricating artificial fingertip sensors with thicker, opaque, soft material removes the superhuman ability to detect these fine features but makes the sensors much easier to fabricate and more robust during use~\cite{Insight}.
Inspired by human touch sensing, which requires both spatial patterns and motion-induced vibrations to perceive texture properties \cite{weberSpatialTemporalCodes2013}, we investigate combining robust moderate-spatial-resolution vision-based touch sensing with the high-frequency modality of audio as a promising approach to fully capture the wide spectrum of touch signals.

To this end, we introduce Minsound, a new soft microphone-based fingertip sensor with a sensing bandwidth of 50~Hz to 15~kHz. The sensor comprises a MEMS microphone enveloped by a soft outer shell, and it demonstrates high sensitivity to vibrations caused by light touch and sliding along materials.
We pair this sensor with the existing vision-based fingertip sensor Minsight~\cite{minsight}, a miniature version of Insight~\cite{Insight}, which senses deformation through a robust, thick silicone layer and therefore cannot perceive textures during static contact. Soft fingertips provide advantages for robotic manipulation such as inherent compliance, higher contact area, translational and rotational friction, and grasp stability. Notably, the sensor shells of our vision-based and audio-based sensors are identical (hollow silicone shell overmolded on an aluminum skeleton), giving us the opportunity to investigate whether the hybrid silicone-aluminum shell, which was optimized to allow the internal camera to measure small contact forces~\cite{Insight,minsight}, also allows for audio-based sensing.

These two sensors are mounted on the thumb and middle finger of a custom robotic hand~\cite{Mack25-ICRA-Estimation}. 
To investigate how each sensing modality influences tactile perception of fabrics, we perform dynamic in-hand gestures on 20 different fabrics.
We fuse the tactile information from both sensors and find that audio sensing performs convincingly well in the challenging task of fabric classification. We can especially show that encapsulating a microphone inside a soft sensor shell captures tactile data more effectively than using only an external microphone, as the internal design achieves a much higher signal-to-noise ratio.
The representations learned with our transformer-based method capture similarities between fabrics and are robust against environmental noise when an external microphone is available.
Our key contributions are:
\begin{itemize}[leftmargin=*]
\item Minsound: A low-cost, soft, audio-based fingertip sensor for high-frequency touch perception;
\item An investigation into the relative importance of vision- and audio-based touch data as well as proprioceptive data for a robot hand using dynamic in-hand exploration to try to recognize fabrics;
\item A multimodal dataset comprising high-resolution images of 20 common fabrics as well as vision- and audio-based touch data and robot proprioception data gathered over about eight minutes of rubbing each fabric. The  full dataset can be downloaded from the \href{https://sites.google.com/view/multimodaltouch}{project website}\footnote{\url{https://sites.google.com/view/multimodaltouch}}.

\end{itemize}

\section{Related Work}
\label{sec:relatedwork}
Dynamic surface recognition with hard tools or hard fingertips has been studied in both haptics and robotics; the most recent research indicates that a high-bandwidth three-axis accelerometer rigidly mounted to the tool is the most promising data source for successful texture recognition \cite{strese2017materialclassification,fuchunsun2020crossmodal,khojasteh2024surfacerecognition}, as it provides richly varying oscillatory data streams that sensitively respond to characteristics of the surface being touched. 
However, using an accelerometer for robotic fingertip exploration comes with the challenge of robustly embedding this rigid component into a soft material. Hard sensors and their wiring harm the compliance and longevity of soft structures, quickly tearing out under load. In contrast to accelerometers, a microphone can be located away from the touch surface, similar to the concept of vision-based tactile sensing, and can capture useful high-frequency temporal information from within a soft enclosure, as shown by Zöller et al.~\cite{brock2018acoustic}.

\subsection{Audio-based Tactile Sensing}
Acoustic tactile sensing is an emerging research direction. Lu et al.~\cite{luActiveAcousticSensing2023a} used a vibration actuator to physically excite rods made of ABS plastic, aluminum, steel, and wood and demonstrated that measuring the propagation and distortion of the waveforms using a piezoelectric microphone enables classification of the rod material. Liu et al.~\cite{liu2024sonicsense} employed contact microphones to discern the filling and material properties of different objects explored with tapping motions. These methods highlight the utility of sound in deducing material characteristics during interactions, but they rely on rigid fingertips or tools for efficient audio propagation from the contact surface to the contact microphone. In contrast, our work uses soft robot fingertips due to their advantages in object manipulation. Wall et al.~\cite{wallPassiveActiveAcoustic2023} recently introduced the approach of sensorizing soft pneumatic actuators with embedded microphones and speakers. This combination enables measurement of contact forces, object materials, and actuator temperature, showcasing the potential of auditory cues in tactile sensing.

\subsection{Fabric Recognition with Robotic Tactile Sensors}
The task of fabric recognition based on robotic tactile sensing has been studied only rarely.
In a static context, Yuan et al.~\cite{yuanActiveClothingMaterial2018b} and more recently Böhm et al.~\cite{B_hm_2024} used vision-based tactile sensing for fabric recognition, applying pressure with a GelSight sensor to capture the small-scale 3D structure of fabric. These approaches deliver impressive recognition results while relying purely on vision-based sensing. However, the data collection process is not human-like, as it does not involve the characteristic excitation of skin vibrations, and it demands the capture of high-resolution images during periods of static pressure.
Ward-Cherrier et al.~\cite{NeuroTac} utilized a unique event-based tactile sensor to recognize textures across 20 different fabrics. Similarly, Cao et al.~\cite{Cao_2020} used Gelsight sensors to explore 100 fabric samples and fuse temporal as well as spatial aspects of the visual data. Both experiments involved fixing fabrics on a rigid board before exploring them through a controlled sliding or pressing motion; with this procedure, the captured signal is somewhat contaminated by the feel of the board. For better generalizability, we aim to explore freely hanging fabrics more as humans do, by rubbing the fabric between two opposing fingers. 

Previous research that uses microphones embedded in a soft structure does not thoroughly investigate fabric recognition. Chang et al.~\cite{changMic} use 10 microphones under a soft, square sensor pad to distinguish four 3D-printed textures. Lambeta et al.~\cite{Digit360} use the audio sensor embedded in the soft fingertip of Digit 360 to distinguish three object materials (plastic, wood and silicone) with less accuracy than reported in this paper. Zöller et al.~\cite{brock2018acoustic} focus on contact localization and report classification accuracy for only three different materials.

\subsection{Multimodal Fusion for Tactile Sensing}
As tactile perception is informed by multiple sensing modalities with distinct characteristics, fusing data from two or more modalities has remained an ongoing challenge. As the first widely distributed multimodal tactile sensor, the BioTac has led to much work on multimodal sensor fusion. A frequent early approach for sensor fusion was to compute hand-crafted features from the various modalities and concatenate them to classify surfaces \cite{fishel2012bayesian}, haptic properties \cite{chu2015robotic}, or objects \cite{hoelscher2015evaluation} or to detect slip \cite{veiga2018grip}. More recent approaches have used learning-based techniques that extract latent representations of each modality before fusing them \cite{richardson2019improving,chen2021tactile}. Yu et al.~\cite{yu2024mimictouch} collect vision-based tactile information and audio from an object-mounted contact microphone during human demonstrations of contact-rich object manipulations. They fuse these modalities with external object-tracking information and use this combination to train successful manipulation policies for a robotic gripper via imitation learning.
The sensor Digit 360~\cite{Digit360} is a promising platform for further research on multimodal tactile fusion; however, its evaluation still lacks experiments to analyze how the different modalities contribute to discrimination of a larger variety of materials.
Most similar to this work, Dhawan et al.~\cite{dhawanDynamicLayerDetection2024} analyze optical flow data from two vision-based tactile sensors rubbing against silk fabric to determine the number of grasped fabric layers. They investigate how flow magnitude, flow angle, contact wrench information, and joint angles from the motors of their custom two-fingered gripper influence their method's performance. Their technique showcases the effectiveness of vision-based tactile sensing combined with proprioception for material assessment but is limited to the evaluation of one kind of fabric.\looseness-1

\section{Method}

To demonstrate and evaluate the proposed dynamic in-hand fabric recognition method,
a modular robot hand is equipped with vision- and audio-based tactile sensors, as described in \cref{sec:hardware}. The fingers perform the data-collection motion that is outlined in \cref{sec:finger_motion} to capture multimodal information about the fabric. We are investigating the realistic scenario of feeling fabrics hanging from a clothes hanger, as one would encounter in a closet or clothing store. In these scenarios, usually more than one layer of fabric is folded between the fingers, which we reproduce in our setup. All sensory modalities and their different representations are introduced in \cref{sec:multimodal_inputs}. Our machine-learning method for fabric classification is outlined in \cref{sec:classification_architecture}.

\begin{figure}[tp]
\begin{center}
\medskip
   \includegraphics[width=\linewidth]{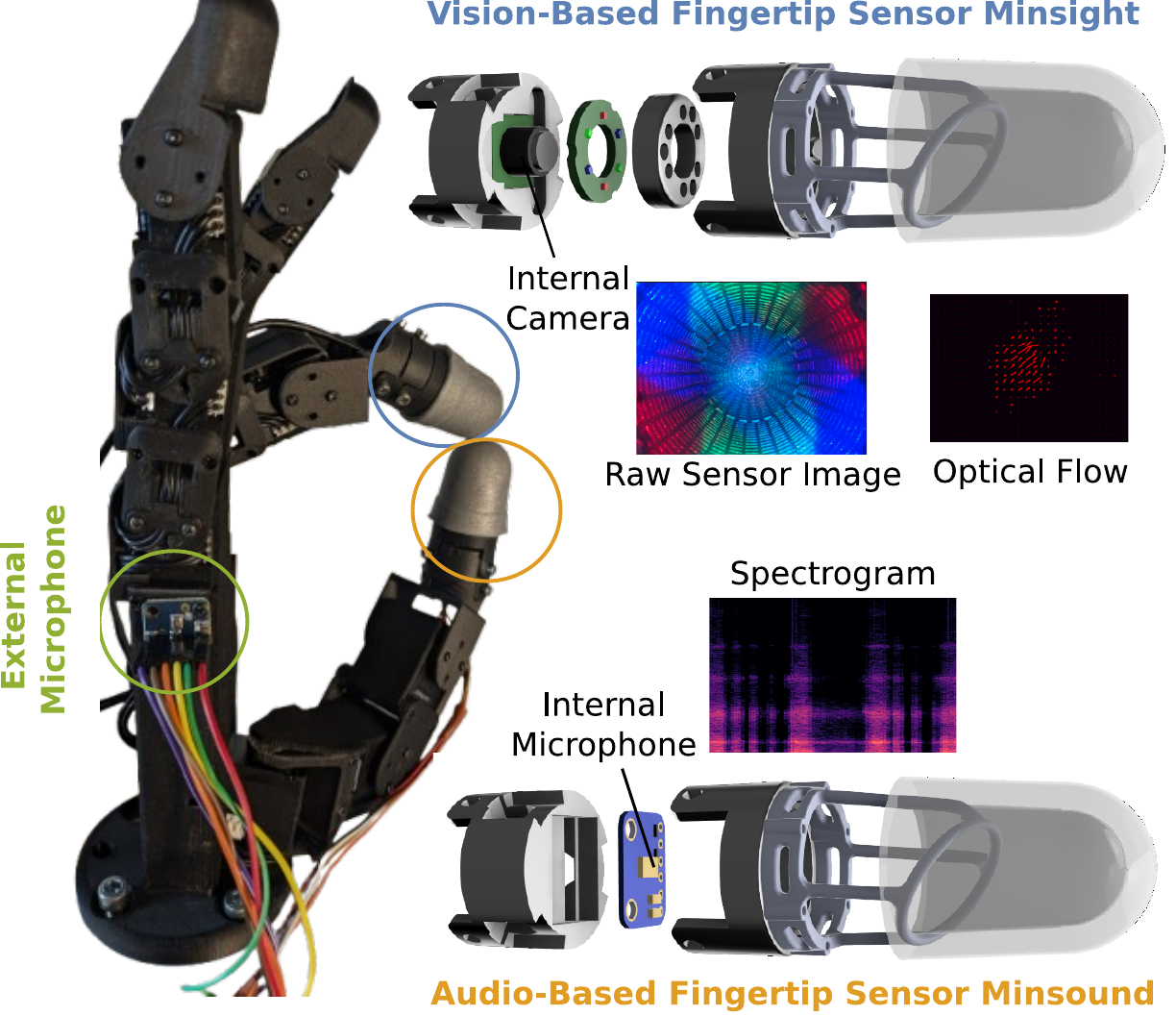}
    \caption{\textbf{Multimodal data collection setup:} A four-fingered robot hand is equipped with two tactile fingertips that measure different modalities of touch. The middle finger terminates with the vision-based tactile sensor Minsight, which delivers high-resolution internal images of the soft fingertip deformation. These images are used to infer a contact force map and to calculate optical flow over consecutive frames. The robot thumb is equipped with Minsound, our soft microphone-based sensor, which records broad-bandwidth audio at a sampling rate of 48\,kHz. An identical microphone is mounted on the side of the robot's palm to record environment noise.}
    \label{fig:hardware_setup}
\end{center}
\end{figure}

\subsection{Hardware Setup}
\label{sec:hardware}
The hardware consists of a low-cost, four-fingered, human-inspired robot hand (ISyHand V2, \cite{Mack25-ICRA-Estimation}) equipped with two fingertip sensors and one external microphone mounted on the side of the palm. The joint angles of the robot hand can be commanded at 50\,Hz.
The hand's middle finger is equipped with an open-source Minsight vision-based tactile sensor \cite{minsight, minsightopensource}, which senses contact in 3D across its entire surface, as shown in \cref{fig:hardware_setup}. It captures images of the textured interior of the opaque silicone rubber shell at a maximum frame rate of 60\,Hz. For our experiments, we sample frames at 50\,Hz for synchronization with the robot hand's joint information.

The robot thumb is equipped with our novel microphone-based tactile sensor Minsound. It is structurally similar to Minsight, i.e., it has the same dimensions and uses the same aluminum skeleton covered in a soft opaque silicone elastomer. Instead of a camera, though, it uses a SPH0645 MEMS microphone as the transducer that is centrally mounted at the base of the sensor, as shown in \cref{fig:hardware_setup}.
This compact microphone fits easily inside the sensor shell and has a bandwidth from 50\,Hz to 15\,kHz. It delivers audio signals to a Raspberry Pi 4 single-board computer at a sampling rate of 48\,kHz with 16-bit precision.
A second SPH0645 microphone is mounted externally to the palm to measure environment noise; its audio signals are processed in the same way.

We selected this specific microphone for Minsound due to its internal analog-to-digital conversion and I2S (Inter-IC Sound) interface for serial communication of the acquired signal. For comparison, we also evaluated the analog MEMS SPW2430 and confirmed that analog microphones are more prone to electrical noise than I2S microphones due to their sensitivity to power supply fluctuations.

\subsection{Finger Movement and Data Collection}
\label{sec:finger_motion}

Given a folded piece of fabric or clothing that is hung from a clothes hanger, the robot hand performs a preprogrammed human-like exploratory procedure \cite{ledermanklatzky} that is designed to use relative lateral motion between the two sensitive fingertips to evaluate the texture of the material. It starts with a force-controlled closing of the thumb and middle finger to grasp the fabric. We use the force-sensing capabilities of Minsight to estimate the current grasping force and stop the closing motion when the force threshold of 0.1\,N is reached, similar to \cite{chu2015robotic}. The closing of the fingers is followed by a preprogrammed rubbing motion that moves the fingers back and forth against the fabric, as shown in the video that accompanies this paper. To facilitate comprehension of this haptic interaction, the video's audio track presents the sound measured by the internal microphone. The closing force of the fingers was optimized to balance smooth rubbing movements and signal generation: the force is set low enough to allow the fingers to rub smoothly without reaching the maximum joint torques and high enough to generate a sound that is detectable by the internal microphone. The same force threshold and commanded finger speed are used for all fabrics. 

\begin{figure}[t]
\begin{center}
\medskip
   \includegraphics[width=\linewidth]{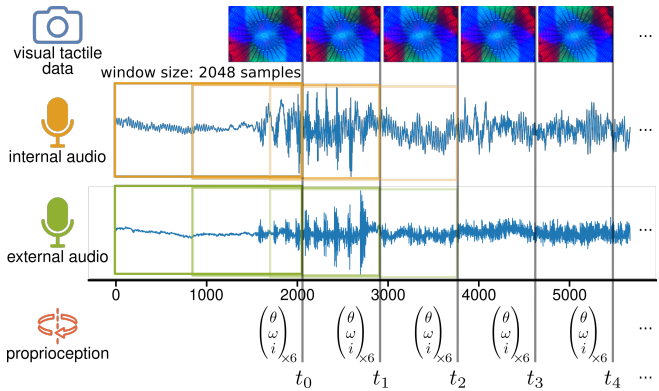}
    \caption{\textbf{Data collection:} During the entire exploratory procedure, we stream camera images from the vision-based tactile sensor at 50\,Hz, audio data from the internal and external microphones at 48\,kHz, and joint angles, currents and velocities for all six involved motors at 50\,Hz (every 20\,ms). For each image and the corresponding proprioceptive data, we record a window of microphone data corresponding to the most recent 2048 audio data points (42.67\,ms) recorded before the image was captured.
    }
    \label{fig:sample_window}
\end{center}
\vspace{-1.0em}
\end{figure}

During the exploratory procedure, we stream data from Minsight's camera, Minsound's microphone, the external microphone, and the hand's joints to the control unit, in this case a standard laptop. For each captured Minsight image, the latest angle, current, and velocity of the six involved joints are saved. Furthermore, for both internal and external microphones, we store a window of 2048 audio data points, corresponding to 42.67\,ms of sound that was recorded immediately before the time of image capture; note that successive audio windows overlap by about 50\% because images are captured approximately every 20\,ms and we want to use all captured audio. The image, both associated windows of audio data, and the vector of joint information constitute a single time sample, as visualized in \cref{fig:sample_window}.

\subsection{Multimodal Inputs}
\label{sec:multimodal_inputs}
To evaluate which data representations are most meaningful for fabric recognition and to analyze how much each haptic modality contributes to the classification performance, we compare different input streams and their representations.

\subsubsection{Minsight Images}
\label{subsec:minsight_images}
The Minsight sensor provides images from its internal camera, which captures the deformation of the soft sensor shell due to external forces. The sensor shell is opaque, so external light does not influence the camera image. This sensor contains colored LEDs that provide internal lighting. We sample images that have a resolution of 308$\times$410 pixels from the Minsight sensor (identified as the original Minsight images in the following) at an average frame rate of 50\,Hz to match the update frequency of the robot hand. Andrussow et al.\ showed that a resolution of 60$\times$80 pixels is sufficient to infer precise forces from these images~\cite{minsight}. We also downsample the images to this resolution to decrease input dimensionality for our experiments.

\subsubsection{Optical Flow}
\label{subsubsec:optical_flow}
Fabric exploration can be approached as a dynamic task, and useful information might lie in the sequential change of deformation, which potentially encodes information about the frictional properties of the contact. We therefore calculate the optical flow between consecutive frames of the raw camera stream for an interaction sequence. We use the Farneback method to calculate dense optical flow over the original Minsight image and then filter for the top 0.1 percentile according to magnitude. We furthermore resize the optical flow image by averaging over patches of 3$\times$3, so the resulting output has dimensions of 2$\times$102$\times$137.

\begin{figure*}%
\begin{center}
\medskip
   \includegraphics[width=0.8\linewidth]{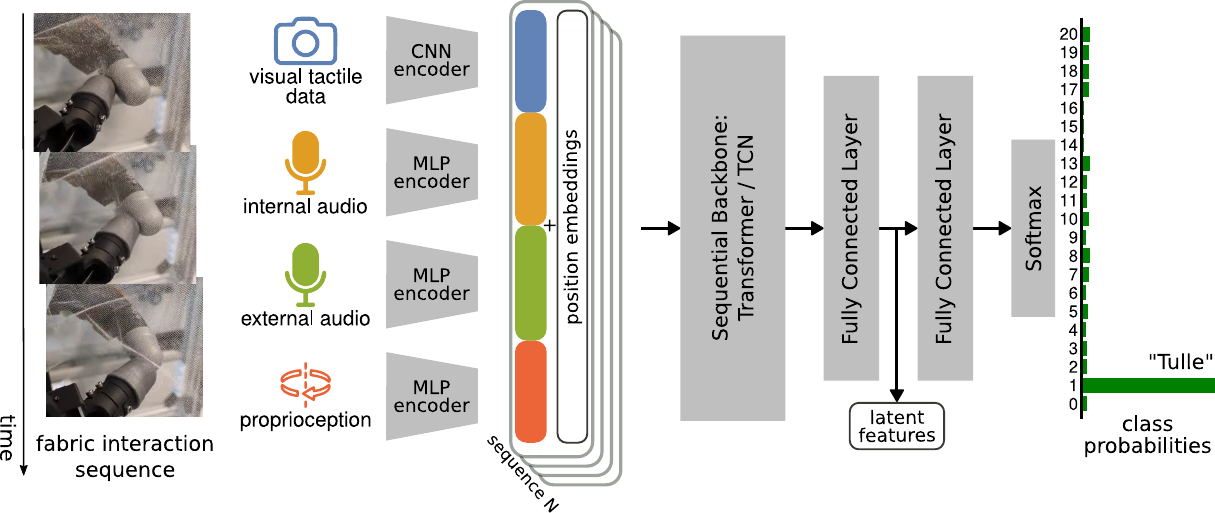}\vspace{-.5em}
    \caption{\textbf{Multimodal classification architecture:} We process fabric interactions in sequences of $N=200$ time steps (4\,s of data). Each modality is processed by an encoder head, and the respective features are concatenated and normalized. A position embedding is added to each overall feature vector, and the sequence of all resulting vectors is processed by a sequential backbone, consisting of three multi-head attention layers or a temporal convolutional network (TCN). The outputs of the backbone are further processed by a classification head that consists of two fully connected layers followed by a softmax function to yield the output.}
    \label{fig:method_overview}
\end{center}
\vspace{-0.5em}
\end{figure*}

\subsubsection{Power Spectral Density}
Both MEMS microphones record at 48\,kHz. Based on preliminary testing against other representations, we use Welch's method to compute the power spectral density (PSD) over the window of length 42.67\,ms before each captured image (2048 data samples, as shown in \cref{fig:sample_window}). The resulting density has 512 frequency bins.

\subsubsection{Proprioception}
The robot hand provides the angle, angular velocity, and current of the servo motor at each joint. As our experiments use only the hand's thumb (three joints), middle finger (two joints) and palm (one joint), the proprioceptive data consists of 6$\times$3 values concatenated to make an 18-dimensional vector.

\begin{figure}[b!]
\begin{center}
   \includegraphics[width=\linewidth]{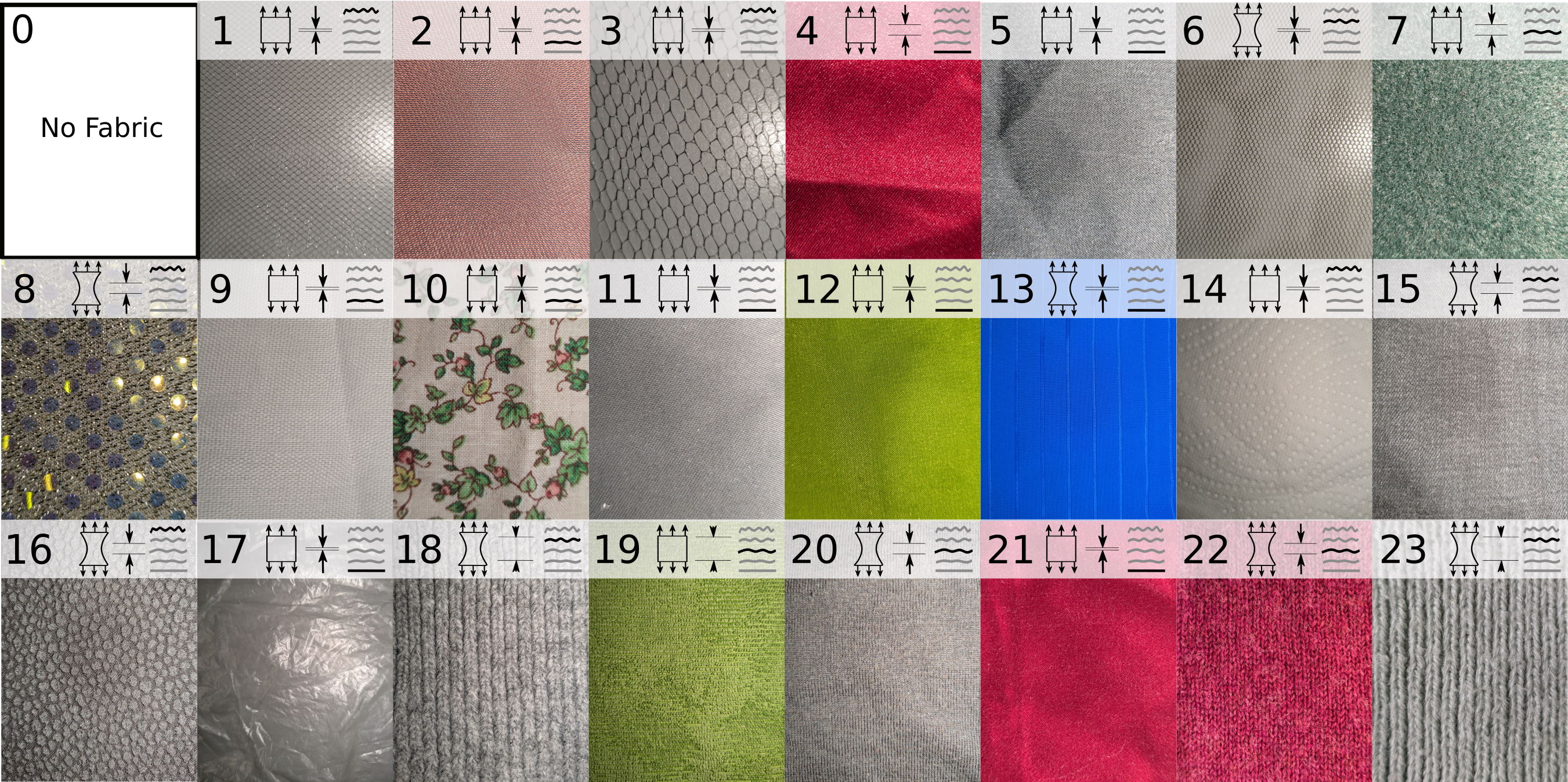}\vspace{-.5em}
    \caption{\textbf{Close-up images of all 20 fabrics used for training plus three holdout fabrics (21--23) to test generalization:} Class 0 is only the robot fingers moving against each other without any fabric. The depicted side of each material is folded inward during data collection. The dataset corresponding to this paper provides verbal descriptions and property categories (stretchiness, thickness and roughness, visualized here by pictograms) of all fabric samples \cite{fabricdataset}.}
    \label{fig:fabrics}
\end{center}
\end{figure}

\subsection{Classification Architecture}
\label{sec:classification_architecture}
As shown in \cref{fig:method_overview}, the proposed method uses a multimodal approach similar to Dhawan et al.~\cite{dhawanDynamicLayerDetection2024} to classify fabrics based on combined visual, audio, and proprioceptive data. Each stream of data is processed through specific encoders: visual data is fed into a convolutional neural network (CNN) encoder, which is co-trained with the backbone, while audio and proprioceptive data are processed through multi-layer perceptron (MLP) encoders. For the co-trained image encoders, we employ data augmentation in the form of random rotations in the range of $-90^\circ$ to $90^\circ$ around the image center to artificially increase the dataset size. The same transformation is applied to all images of a sequence.

The weights of the encoders for both audio input streams are shared. The resulting features of all encoders are concatenated into a combined feature vector and then stacked in time to a sequence of length $N=200$, which corresponds to 4\,s of interaction, the duration that was found to be optimal in preliminary experiments. To remove magnitude dependencies, each dimension of the combined feature vector is rescaled to $[0,1]$ by normalizing over the time sequence. The time sequence is then processed using multi-head attention, allowing the model to weigh the importance of different features during classification. After attention processing, the combined features are sent to a classification head consisting of two linear layers for the final fabric-type identification. For detailed hyperparameters, please refer to the code published on our project website. Inference with the transformer model over a sequence of 200 samples takes 5.58\,ms on an A100 GPU.

\section{Experiments}
This section investigates the following research questions:
\begin{enumerate}
\item Can a robot hand equipped with soft vision- and audio-based tactile fingertip sensors distinguish between different fabrics using a dynamic exploratory procedure?
\item How do the tested haptic sensory modalities (vision, audio, proprioception) contribute to fabric recognition?
\item How robust to environmental noise is fabric classification using audio-based tactile sensing?
\item Can the proposed architecture be used to infer properties of fabrics that were not previously touched?
\end{enumerate}

To evaluate our method, we collected a dataset of fingertip interactions with a set of 20 diverse fabrics \cite{fabricdataset}. These fabrics include tailoring materials such as silk and tulle as well as everyday items such as a knitted wool hat, an old microfiber towel, and a swimsuit. All fabrics are depicted in \cref{fig:fabrics} and are further described in the dataset; interactions with no fabric (class 0) were included for comparison. Each fabric sample was hung from a clothes hanger above the hand, as shown in \cref{fig:overview}. We collected interactions on different days to capture variations in environmental noise and to prevent spurious correlations in the sound profile from influencing the recognition performance. The data used in the training set were collected on two days; on each day, the robot performed six exploration trials in a row for each fabric under ambient lab environmental noise. After each trial, the position of the clothes hanger was adjusted to create variation in the grasping location on the fabric. Testing data were collected on a separate day with only two successive trials per fabric. On all days, fabrics were presented in random order. In total, there are twelve training trials and two testing trials per label.

\begin{figure}
\begin{center}
\medskip
   \includegraphics[width=\linewidth]{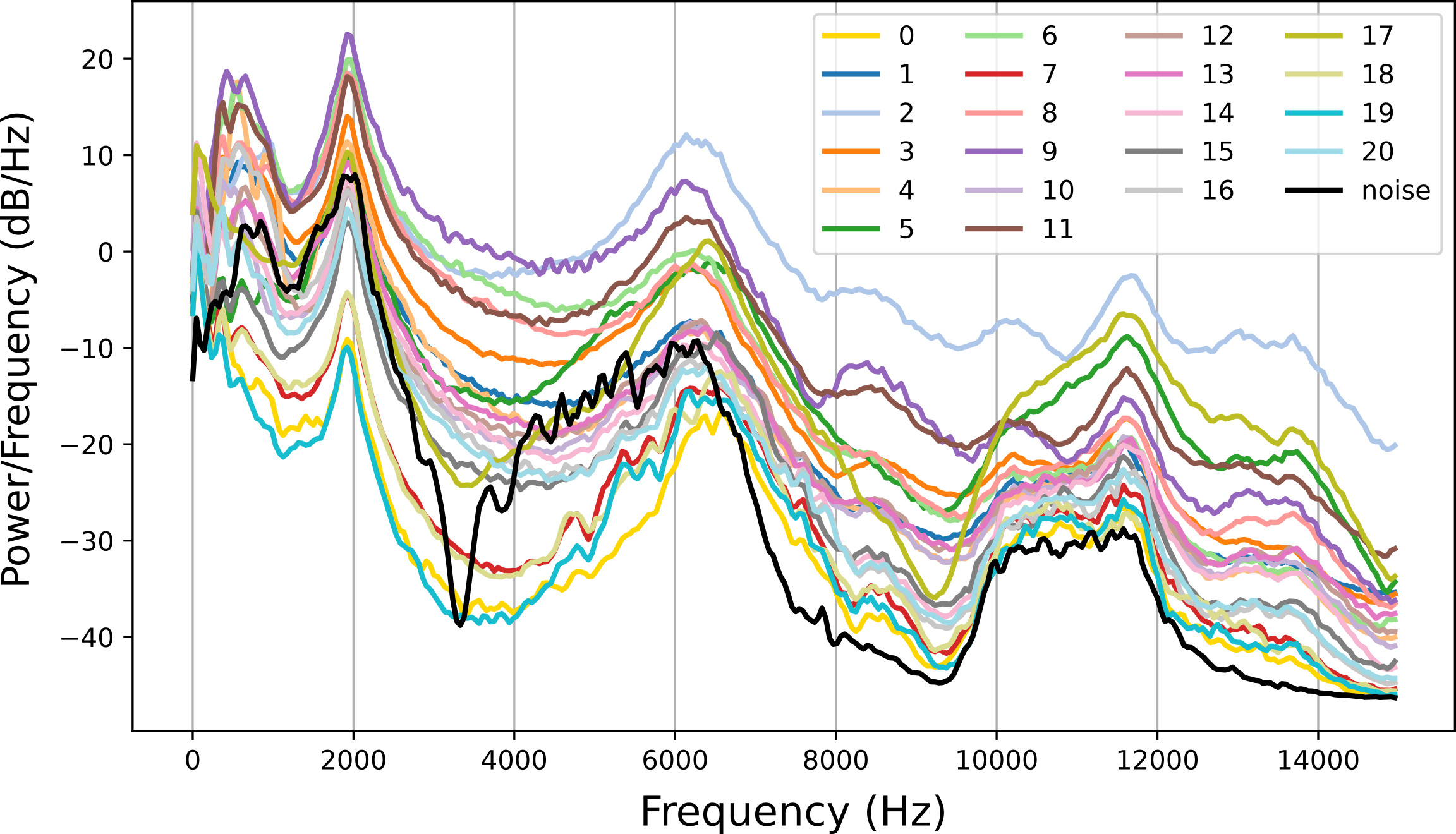}
    \caption{\textbf{Mean power spectral densities of all fabric classes:} Examining the mean power spectral density over one trial for each label in the dataset shows that the identity of the fabric being touched seems to strongly affect the recorded tactile sound. No fabric (class 0) has the smallest signal at most frequencies. Minsound's measurement of the noise signal used in \cref{sub:noise_results} appears in black.}
    \label{fig:psds}
\end{center}
\end{figure}

Each trial contains 50 individual finger rubbing motions, during which measurements from all modalities are continuously recorded. This procedure results in around 1500 individual time steps per trial, which differs across trials, as the time for execution of 50 rubbing motions depends slightly on the type of fabric: the fingers move more slowly when finger-fabric friction is high. The trials are partitioned into non-overlapping sequences of 200 time steps (4\,s at 50\,Hz), and these sequences are used as individual training samples, resulting in 1588 training samples and 273 testing samples over 21 classes. The mean power spectral densities recorded by Minsound over all samples of one trial for each label are shown in \cref{fig:psds}. We evaluate our model based on different input features and backbone architectures in \cref{subsec_ablation}. 

To assess our approach's sensitivity to background noise, we record another interaction dataset for all fabrics while playing audible noise. The performance of our method on this test dataset is presented in  \cref{sub:noise_results}.
\cref{sec:latents} analyzes our method's ability to infer common fabric properties for previously untouched fabrics and evaluates the latent representations learned.

\begin{table}
    \centering
    \vspace{2ex}
    \caption{Validation and test accuracy for four audio bandwidths.}
    \begin{tabular}{ccc} 
        \toprule
        Bandwidth & Validation Accuracy & Test Accuracy  \\      
        
        [kHz] & [\%] & [\%] \\
        \midrule
        2.5 & 92.43 $\pm$ 1.13  & 89.87 $\pm$ 3.31  \\ %
        10 & 97.45 $\pm$ 1.17 & 93.07 $\pm$ 3.33  \\ 
        15 & 97.75 $\pm$ 1.62 & \textbf{95.06 $\pm$ 1.85}  \\ 
        24 & 98.20 $\pm$ 1.39 & 91.94 $\pm$ 2.68  \\ 
        \bottomrule
    \end{tabular}
    \label{tab:bandwidth_classification}
\end{table}

\newcommand{\cw}{@{\hspace{5pt}}}
\begin{table}%
    \centering
    \caption{Model performance on different input modalities.}\vspace{-1em}
    All results are reported as mean $\pm$ SD for five seeds.\vspace{.3em}
    
    \adjustbox{max width=\linewidth}{
    \begin{tabular}{@{}l@{\,}c\cw c\cw c\cw c\cw c\cw c@{}}
        \toprule
        Backbone & Minsight &  Optical & Minsound & External & Proprio- & Test \\
         &  Image &   Flow &  Audio & Audio & ception &  Accuracy [\%] \\ \midrule

        Transformer         & $\checkmark$   &                &                               &                    &                  & 41.15 $\pm$ 5.74\\
        Transformer         &                &   $\checkmark$ &                               &                    &                  & 47.16 $\pm$ 5.12\\
        Transformer         &                &                &         $\checkmark$          &                    &                  & \textbf{95.06 $\pm$ 1.85}\\
        Transformer         &                &                &                               &   $\checkmark$     &                  & 59.22 $\pm$ 2.43\\
        Transformer         &                &                &                               &                    &   $\checkmark$   & 47.10 $\pm$ 3.72\\
        \midrule
        Transformer         &  $\checkmark$  &                &          $\checkmark$         &                    &                  & 73.94 $\pm$ 5.73\\
        Transformer         &  $\checkmark$  &                &          $\checkmark$         &   $\checkmark$     &   $\checkmark$   & 85.59 $\pm$ 3.87\\
        
        Transformer         &                &   $\checkmark$ &          $\checkmark$         &                    &                  & 87.44 $\pm$ 5.88\\
        Transformer         &                &   $\checkmark$ &          $\checkmark$         &   $\checkmark$     &   $\checkmark$   & 91.34 $\pm$ 2.47\\

        Transformer         &                &                &          $\checkmark$         &                    &   $\checkmark$   & 94.63 $\pm$ 2.78\\
        Transformer         &                &                &          $\checkmark$         &   $\checkmark$     &   $\checkmark$   & 96.10 $\pm$ 1.26\\
        Transformer         &                &                &          $\checkmark$         & $\checkmark$       &                  & \textbf{97.75 $\pm$ 1.27}\\
        \midrule
        
        TCN                 &                &                &          $\checkmark$         &   $\checkmark$     &   $\checkmark$   & 75.41 $\pm$ 11.2\\ \bottomrule
    \end{tabular}
    }
    \label{tab:ablation_table}
    \vspace{-1ex}
\end{table}

\subsection{Frequency Spectrum Analysis}

The microphone has a specified bandwidth of 50\,Hz to 15\,kHz, but audio can be sampled at 48\,kHz, which makes a frequency analysis up to 24\,kHz possible. To analyze whether high-frequency components above the specified bandwidth of 15\,kHz are useful for fabric classification, we train the classification pipeline with different bandwidths by truncating the PSD at the respective maximum frequency; the results appear in \cref{tab:bandwidth_classification}. A bandwidth going beyond 15\,kHz causes the model to overfit, while a bandwidth lower than 10\,kHz reduces the test accuracy. We thus use 15\,kHz.

\subsection{Input Feature Ablation}
\label{subsec_ablation}
\begin{figure}
\smallskip
\begin{center}
   \includegraphics[width=\linewidth, trim=0 0 0 0.5cm, clip]{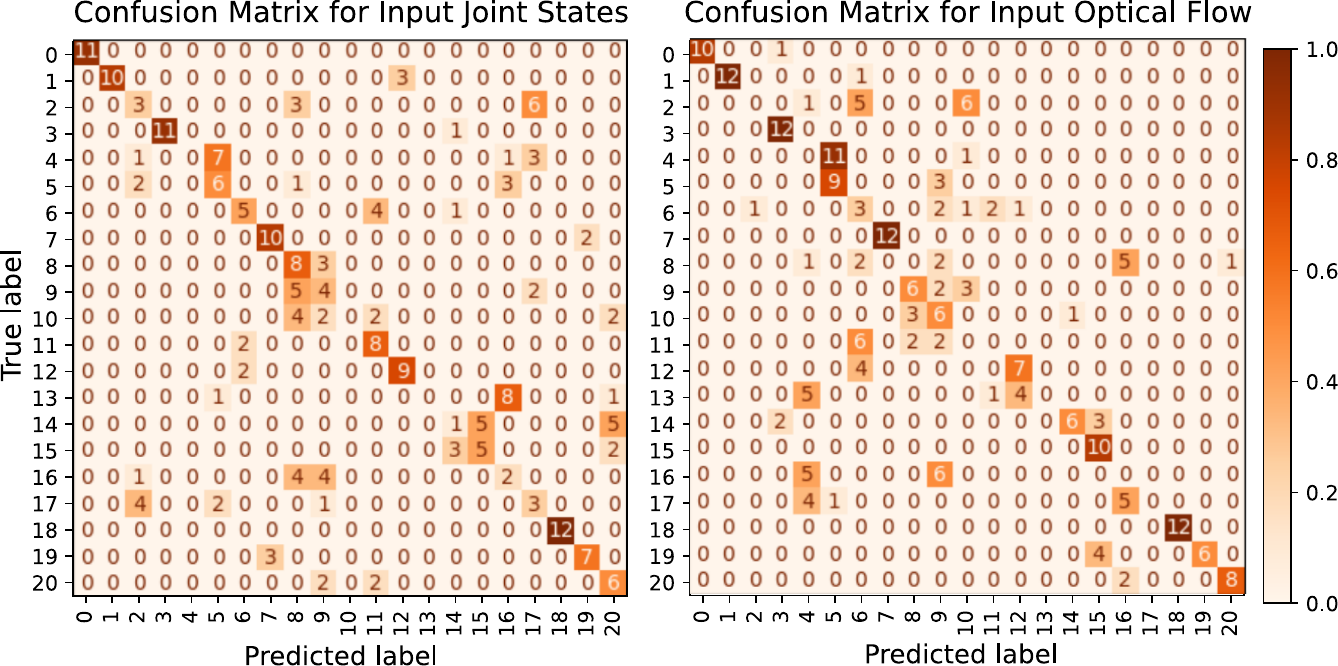}
    \caption{\textbf{Confusion matrices on the test set for classification using only proprioceptive information (left) or only optical flow (right).}}
    \label{fig:confusion_matrix}
\end{center}
\vspace{-2ex}
\end{figure}

We conducted experiments to analyze the influence of each input modality on the overall classification accuracy. 
The top part of \Cref{tab:ablation_table} shows that Minsound's internal microphone outperforms all other single sensing sources by a large margin, achieving a mean classification accuracy of 95.06\% on the test set. This result demonstrates that our soft audio-based fingertip sensor is capable of distinguishing a wide variety of fabrics. Using Minsight's images, on the other hand, leads to overfitting of the network and, therefore, to poor generalization to the test set (41.15\% accuracy).
We additionally test a more task-specific representation of the raw images, namely optical flow calculated over the image sequence, as described in \cref{subsubsec:optical_flow}. As shown in \cref{fig:confusion_matrix}, using solely proprioceptive information (47.10\%) produces a confusion matrix that is similar along the diagonal to that obtained using solely optical flow information (47.16\%), which indicates that these two modalities can be used to distinguish similar kinds of fabrics.

As shown in the bottom part of \Cref{tab:ablation_table}, adding the external audio improves classification accuracy by two percentage points to the best overall performance of 97.75\%. This value is significantly higher than that achieved with internal audio alone, as found by a Wilcoxon signed-rank test ($p = 0.0254$).

\subsection{Robustness to Environmental Noise}
\label{sub:noise_results}

To assess the robustness of our approach against environmental noise, we evaluate the fabric classification on an additional test set that was recorded with loud background noise resembling ambient sound in a coffee shop. The average PSD over one minute of this noise is shown in black in \cref{fig:psds}. While the training data were captured in a laboratory setting where the noise level near the external microphone was measured to be in the range of 29 to 31\,dB, the coffee shop noise lies in a range of 47 to 55\,dB, as measured with a Sauter SU 130 sound meter. The results in \cref{tab:new_noise_results} show that using only Minsound's internal microphone for this task decreases classification accuracy to 67.51\%. Adding proprioceptive information increases the accuracy to 87.51\%, which further increases to 91.57\% when including the external microphone. The table reports statistical test results that confirm the significance of these improvements.

\begin{table}[t]
    \medskip
     \caption{Classification results under noisy environmental conditions.}\vspace{-1em}
        Accuracies are mean $\pm$ SD over 10 seeds. Each $p$-value is reported for a Wilcoxon signed-rank test with the preceding row.
     \begin{center}
    \begin{tabular}{ccccc}
        \toprule
        Minsound & Proprio- &  External& Test Accuracy & $p$-value \\
        Audio & ception &  Audio& [\%] &  \\
        \midrule
        $\checkmark$ &              &  &  67.51 $\pm$ 5.05  & -- \\
        $\checkmark$ & $\checkmark$ &  &  87.51 $\pm$ 2.66  & < 0.0001\\
        $\checkmark$ & $\checkmark$ & $\checkmark$ & \textbf{90.30 $\pm$ 2.18} & \phantom{<} 0.0098 \\
        \bottomrule
    \end{tabular}
    \end{center}
    \label{tab:new_noise_results}
\end{table}

\subsection{Property Perception and Learned Representations}
\label{sec:latents}

To analyze the generalization capabilities of our method beyond recognition of fabrics included in the training dataset, we train the architecture to classify three properties that humans commonly use to describe fabrics (two levels of stretchiness, five levels of roughness, and three levels of thickness). We evaluate the performance on two fabrics held out from the training set (1 and 10) as well as three new fabrics: thin real silk (21), a wool sweater (22, shown in \cref{fig:overview}), and a knitted acrylic hat (23). Our dataset \cite{fabricdataset} provides descriptions and thickness measurements for these additional fabrics, plus the property categories for all fabrics. 

For each property, a different classification head is learned, which replaces the last layer of the architecture shown in \cref{fig:method_overview}. The rest of the architecture is shared across all properties. All heads are trained jointly with the base model by adding all classification losses with equal weighting. \cref{tab:property_perception} shows the results of the property-perception experiments for each property on the 18 fabrics contained in this training set and on the five holdout fabrics, as well as a comparison between using only Minsound or Minsound~+~Minsight as sensing modalities. Proprioceptive information is used across all property-perception experiments, and the visual input is in the format of downsampled Minsight images. All three properties are perceived with accuracies higher than chance on the holdout fabrics. Minsound~+~Minsight outperform audio alone when estimating stretchiness and thickness, while roughness is perceived better by Minsound alone.

\begin{table}[tb]
\medskip
    \caption{Property perception.}\vspace{-1em}
    Mean accuracy [\%] for property classification across 5 seeds. For this task, the five holdout fabrics are numbers 1, 10, 21, 22, and 23 in \cref{fig:fabrics}
    \begin{center}
    \begin{tabular}{@{}l\cw c\cw c\cw c@{}c \cw c\cw c@{}}
        \toprule
        &  & \multicolumn{2}{c@{}}{Training Fabrics}&& \multicolumn{2}{@{}c}{Holdout Fabrics}\\\cmidrule{3-4}\cmidrule{6-7}     
                     & Chance & Minsound & {}Minsound && Minsound & {}Minsound \\ 
                     &        &          &+Minsight &&          & +Minsight         \\
                     \midrule
        Stretchiness & 50.00   & 97.40 & 85.99     && 76.15 & 82.95    \\ 
        Roughness    & 20.00   & 97.47 & 82.38     && 40.23 & 33.94   \\ 
        Thickness    & 33.33   & 97.59 & 90.43     && 70.56 & 90.47    \\
        \bottomrule
    \end{tabular}
    \end{center}
    \label{tab:property_perception}
    \vspace{-1.0em}
\end{table}

\begin{figure}[t]
\begin{center}
    \medskip
   \includegraphics[width=\linewidth]{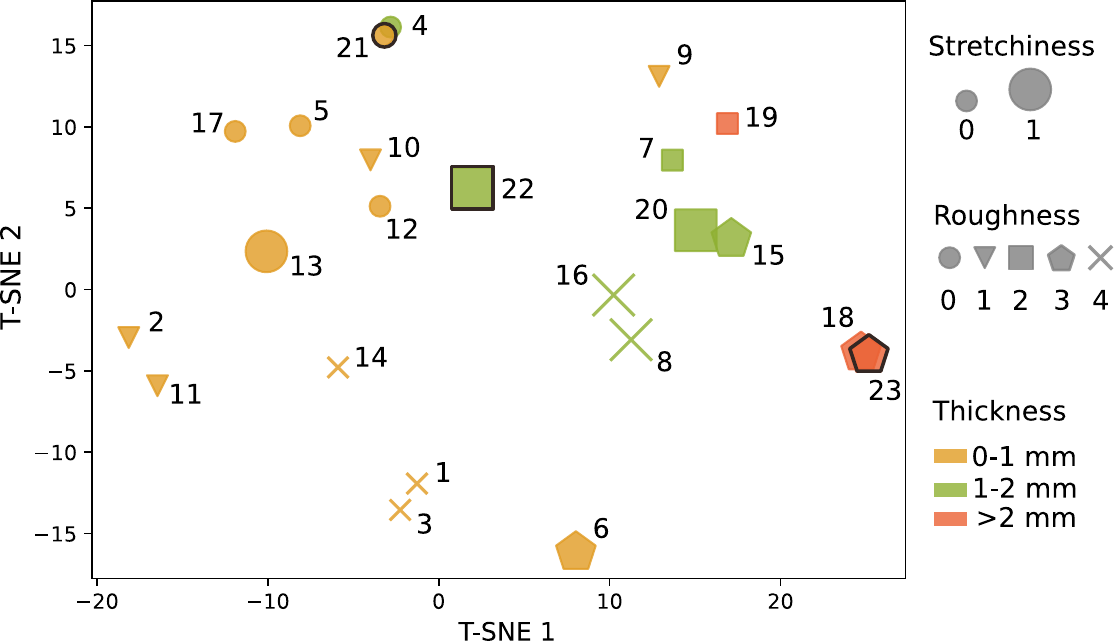}
    \caption{\textbf{T-SNE plot of latent features:} The best-performing model using all modalities creates a latent feature space that shows similarities between fabrics. The mean over inputs from one trial is plotted for each fabric in the training set (1--20). The mean of the test trial for each of the three holdout fabrics (21, 22, 23) is indicated with a black outline. Other marker properties show fabric stretchiness, roughness, and thickness.} %
    \label{fig:latent_features}
\end{center}
\vspace{-1.0em}
\end{figure}

To analyze the representations learned by the property-classification architecture, we plot the mean of the latent features across all modalities for one trial per fabric in \cref{fig:latent_features}. Our analysis shows that the embeddings of different fabrics with similarities across all properties are close together. Furthermore, holdout fabrics are embedded close to similar fabrics within the dataset, meaning that the learned representations can partially capture the space of fabrics. It is notable that the holdout real silk (21) was embedded close to a smooth synthetic fabric (4) but not close to the previously touched real silk (12) from the training set, potentially because they are woven differently.

\subsection{Limitations}
A few limitations should be considered when interpreting these results. The exploratory finger movement is preprogrammed and does not explore the fabric in all directions. Thus, our method might have difficulty learning from any anisotropy in fabrics. Furthermore, the exploratory procedure requires manual tuning of the force threshold and finger speed. Teleoperation by a human would likely generate more diverse data, but such an approach is challenging to implement in practice because of the subtle contact-rich motions that occur during fabric interaction; vision-based hand trackers and tactile gloves greatly struggle to capture such movements. Due to the time-consuming nature of data collection, we use a limited set of fabrics, which is not necessarily representative of the variety of existing fabrics. Nonetheless, we hope that our strong findings motivate further investigation of audio-based soft tactile sensing, multimodal processing, and the perceptual task of in-hand fabric recognition.

\section{Conclusion}
Our approach uses a multimodal framework to combine the open-source vision-based tactile sensor Minsight with Minsound, our new low-cost audio-based tactile sensor. This work is a proof of concept that showcases the complementary sensing capabilities of internal cameras and internal microphones for soft robot fingertips. We thus plan to integrate both vision-based and audio-based sensing into a single soft fingertip sensor with the same size as Minsight.

We observe that audio sensing significantly enhances the robotic perception of fabrics, especially when the microphone is located close to the contact location within a soft fingertip shell.
Furthermore, we find that using both internal and external microphones to capture audio during fabric interaction improves generalization in the presence of loud, irregular environmental noise.
Notably, using spatial information from the Minsight sensor, even for downsampled images, leads to overfitting on our small dataset, which contains only roughly 80 samples for each class. However, the calibrated vision-based sensor ensures a constant normal force across fabrics and is therefore essential for our robotic setup even if it does not improve fabric classification over audio-based tactile sensing. The generalization advantage of audio-based tactile sensing from small datasets may be especially valuable in robotics where acquiring real interaction data is costly and interactions between soft fingertips and textured fabrics are difficult to simulate realistically.

\section*{Acknowledgments}
This work was supported by the Max Planck Society and the Robotics Institute Germany (RIG), which is funded by the German Federal Ministry of Research, Technology and Space (BMFTR) under grant number 16ME1008. The authors thank the International Max Planck Research School for Intelligent Systems (IMPRS-IS) for supporting I.A.; the Computation and Cognition Tübingen Summer (CaCTüS) Internship for supporting J.S.; Bernard Javot, Felix Grüninger, and others in the MPI-IS Robotics ZWE for technical support; and Andrew Schulz for support on visualizations and video editing.

\bibliographystyle{IEEEtran}
\bibliography{references}

\end{document}

%% file: header.tex
\usepackage{times}
\usepackage[utf8]{inputenc} %
\usepackage[T1]{fontenc}    %
\usepackage{url}            %
\usepackage{booktabs}       %
\usepackage{nicefrac}       %
\usepackage{microtype}      %
\usepackage{adjustbox}      %
\usepackage{graphics,color}
\usepackage{wrapfig}

\usepackage{algorithm}
\usepackage[noend]{algpseudocode}
\algrenewcommand\algorithmicindent{1em}
\algrenewcommand{\algorithmiccomment}[1]{%
\bgroup\hskip2em\textcolor{ourdarkgreen}{//~\textsl{#1}}\egroup}

\let\labelindent\relax %
\usepackage{enumitem}

\usepackage{amsfonts}       %
\usepackage{amsmath}       %
\usepackage{amssymb}
\usepackage{xspace}

\usepackage{xr-hyper}
\usepackage{hyperref}%
\usepackage[capitalise]{cleveref} %

\makeatletter
\DeclareRobustCommand\onedot{\futurelet\@let@token\@onedot}
\def\@onedot{\ifx\@let@token.\else.\null\fi\xspace}
\makeatother

\makeatletter
\newcommand*{\addFileDependency}[1]{%
  \typeout{(#1)}
  \@addtofilelist{#1}
  \IfFileExists{#1}{}{\typeout{No file #1.}}
}
\makeatother

\usepackage{xcolor}
\definecolor{ourblue}{rgb}{0.368,0.507,0.71}
\definecolor{ourorange}{rgb}{0.881,0.611,0.142}
\definecolor{ourgreen}{rgb}{0.56,0.692,0.195}
\definecolor{ourred}{rgb}{0.923,0.386,0.209}
\definecolor{ourviolet}{rgb}{0.528,0.471,0.701}
\definecolor{ourbrown}{rgb}{0.772,0.432,0.102}
\definecolor{ourlightblue}{rgb}{0.364,0.619,0.782}
\definecolor{ourdarkgreen}{rgb}{0.572,0.586,0.}

\definecolor{ourcyan2}{rgb}{0.125,0.722,0.804}
\definecolor{ourred2}{rgb}{0.863,0.184,0.047}
\definecolor{ouryellow2}{cmyk}{0,0.16,1.0,0.07}
\definecolor{ourviolet2}{cmyk}{0.55,0.56,0,0.47}
\definecolor{ourorange2}{cmyk}{0,0.46,0.89,0.11}